# EFT-CoT: A Multi-Agent Chain-of-Thought Framework for Emotion-Focused Therapy

**Lanqing Du[1] Yunong Li[2] Yujie Long[1] Shihong Chen[2*]**

[1]School of Computer Science, Guangdong University of Foreign Studies South China Business College

[2]School of Information Science and Technology, Guangdong University of Foreign Studies

## Abstract

The use of large language models (LLMs) for Mental Health Question Answering (MHQA) has emerged as a promising approach to mitigating shortages in mental health resources. While prior work has strengthened the logical structure of interventions by incorporating Cognitive Behavioral Therapy (CBT), most approaches remain predominantly “top-down,” emphasizing rational cognitive restructuring. This provides limited support for help-seekers’ embodied experience and the processing of primary emotions, making it difficult to address complex emotional distress. To address this gap, we propose EFT-CoT, a multi-agent chain-of-thought framework grounded in Emotion-Focused Therapy (EFT), which adopts a “bottom-up” strategy and operationalizes the intervention process as a three-stage reasoning workflow: Embodied Perception—Cognitive Exploration—Narrative Intervention. The framework consists of eight specialized agents that collaboratively cover key processes such as somatic awareness mapping, adaptive evaluation, core belief extraction, and narrative restructuring. Based on this framework, we perform process-level augmentation on approximately 67,000 real help-seeking texts to construct a high-quality instruction-tuning dataset, EFT-Instruct, and further fine-tune a dedicated model, EFT-LLM. Experimental results show that EFT-LLM significantly outperforms multiple strong baselines and human responses on dimensions such as empathic depth and structural professionalism. Ablation experiments further validate the contributions of key mechanisms within EFT-CoT. Additionally, the white-box audit indicates high consistency and traceability of intermediate states at critical stages. Overall, this

---

[*] Corresponding author.

Email-adress:201104@gwng.edu.cn,csh@gudfs.edu.cn (Lanqing Du), 20241050037@mail.gdufs.edu.cn (Yunong Li), 2520601158@e.gwng.edu.cn (Yujie Long), csh@gdufs.edu.cn (Shihong Chen).

study provides a reproducible implementation path and empirical support for embedding EFT intervention mechanisms into large language models via a "framework–data–model" pipeline.

## 1. Introduction

In recent years, the rapid development of large language models (LLMs) has introduced new possibilities for automated psychological support (Qiu & Lan, 2025; Xie et al., 2025). In online communities, many help-seekers express emotional distress in text, making mental health question answering (MHQA) a low-threshold form of support. MHQA often requires integrating contextual understanding with supportive strategies within a single-turn response. When faced with information-dense and emotionally complex help-seeking texts, the system must not only understand the context of the events but also identify multi-layered emotions and compose responses that integrate empathy, explanation, and guidance. This requires strong semantic understanding as well as rigorous logical organization (Sun et al., 2021).

Benefiting from the generative abilities of LLMs, recent research has begun to explore incorporating Cognitive Behavioral Therapy (CBT) into MHQA (David et al., 2018; Hofmann et al., 2012). Existing work generally falls into two categories: one category drives LLMs to generate supportive responses end-to-end through prompt engineering (Chen et al., 2023; Qiu et al., 2024); the other processes and models key steps of CBT, identifying users' cognitive distortions and guiding cognitive restructuring (Chen & Liu, 2025; Na, 2024). Although these methods have improved response effectiveness to some extent, CBT often follows a "top-down" cognitive processing paradigm, emphasizing rational appraisal to reframe emotional experiences. Psychological research indicates that many help-seekers are not lacking in rational judgment but are instead overwhelmed by unprocessed deep emotional pain. For such individuals, purely cognitive correction often proves ineffective and may even be perceived as preachy due to the neglect of their present emotional experience (David et al., 2018; Etkin et al., 2015; Hofmann et al., 2012; Ochsner & Gross, 2005). In contrast, Emotion-Focused Therapy (EFT) advocates for a "bottom-up" approach, following the intervention logic of "you cannot leave a place until you have truly arrived there": only by fully accessing primary emotions and connecting with their embodied experience can emotional transformation be supported (Greenberg, 2004; Pascual-Leone & Greenberg, 2007). To illustrate this difference, Fig. 1. presents a comparison of two generative paradigms using the same help-seeking text. CBT-style responses emphasize cognitive reappraisal and correction and, in this case, may be perceived as preachy; by contrast, EFT-style responses prioritize acceptance and proximity to the help-seeker's present experience and self-protective reactions, thereby supporting subsequent emotional processing and transformation. Notably, this paper does not aim to determine whether "LLMs + CBT" or "LLMs + EFT" is superior; instead, we focus on computationally modeling EFT's core intervention logic and using it to constrain MHQA generation. This approach aims to improve empathic alignment while enhancing the traceability and auditability of the generation process. Despite the potential value of integrating EFT into MHQA, computational modeling of EFT still faces three technical challenges:

(1) Emotional Level Misalignment: General models struggle to penetrate the surface-level secondary emotions (e.g., irritation) of help-seekers and accurately identify the underlying primary emotions (e.g., helplessness), which may lead to a misalignment of the empathic target.

(2) Lack of Embodied Perception: Existing systems lack explicit modeling of bodily experiences, making it difficult to extract embodied cues from the text context (Schmidmaier et al., 2024). Moreover, generating embodied metaphors that align with real experiences to foster deep empathy is challenging (Damasio, 1996), and responses often remain at the logical level of generalized comfort (Chung et al., 2023).

(3) Limited Interpretability of Intervention Logic: Emotional transformation requires rigorous reasoning, but end-to-end models typically operate as black-box systems, lacking explicit constraints and auditability of intermediate stages (Na, 2024; Yang et al., 2024). As a result, it is difficult to verify whether the model adheres to the predefined EFT pathway, and it is also challenging to evaluate mechanisms and attribute errors when the model fails. Although sophisticated prompting can improve surface-level outputs, it does not resolve this fundamental issue of lacking explicit intermediate constraints and auditability.

To address the above challenges, as shown in Fig. 2, we propose EFT-CoT, which operationalizes EFT intervention logic as a three-stage reasoning process—Embodied Perception, Cognitive Exploration, and Narrative Intervention—and produces traceable, structured intermediate states. To instantiate this reasoning flow, EFT-CoT adopts a multi-agent collaborative design and decomposes the intervention process into eight specialized agents: early-stage agents focus on emotional hierarchy identification and somatic awareness mapping, mid-stage agents focus on adaptive evaluation, core belief extraction, and psychological needs identification, and late-stage agents focus on narrative restructuring and constraint-based response generation. Outputs from each stage are used not only to generate the final response but also to support subsequent process auditing and error analysis. We emphasize that this paper focuses on the computational modeling and mechanism verification of therapeutic processes, and we do not advocate deploying the proposed method as a substitute for clinical care.

To validate the framework and support resource-constrained local deployment scenarios, we first use EFT-CoT to drive large models to perform process-level augmentation on real-world consultation data, yielding a high-quality instruction-tuning dataset. We then fine-tune an open-source model on this dataset to obtain EFT-LLM, which learns EFT-aligned intervention patterns without requiring explicit chain-of-thought disclosure at inference time.

The main contributions of this paper are summarized as follows:

(1) To the best of our knowledge, this is the first work to computationally model an EFT intervention pathway for MHQA. We propose EFT-CoT, which improves a model's ability to perceive and process complex emotions by operationalizing an EFT-consistent mechanism of Embodied Perception, Cognitive Exploration, and Narrative Intervention. Functional ablation and process-level auditing show that the gains primarily arise from the effective implementation of key EFT mechanisms.

(2) Using EFT-CoT, we perform process-level augmentation on approximately 67,000 real help-seeking texts and construct the instruction dataset EFT-Instruct. Unlike conventional datasets that contain only question–answer pairs, EFT-Instruct externalizes expert-guided intermediate process signals as learnable supervision, mitigating the difficulty of acquiring deep intervention logic in the absence of process supervision.

(3) We fine-tune EFT-LLM on EFT-Instruct and conduct systematic evaluations. EFT-LLM outperforms strong baselines in multi-dimensional professional assessments, indicating that EFT-Instruct transfers effectively to open-source models and yields more robust psychological support while balancing reasoning efficiency and deployment cost.

| **User:** "I feel chest pain and my hands are numb... I feel like I'm about to turn into a pile of mud collapsing on the ground. I feel disconnected, floating behind my body... I'm just numb and broken. " | **[ Key Psychological Markers ]**<br>🚩 Somatic:Chest pain, Numbness<br>🧩 Metaphor: "Pile of Mud"<br>⚠️ State: High Arousal / Exhaustion |
|---|---|
| **Paradigm A:** CBT: Reframe + Ground 🧠 **(Correction)**<br>Agent: "It sounds like you are experiencing severe anxiety. However, thinking you are 'turning into mud' is a Cognitive Distortion known as 'Catastrophizing'. You are physically safe. Please use the '5-4-3-2-1 Grounding Technique' to pull yourself back to reality." | 😨 User: "He thinks I'm irrational. I don't have the energy to count things. I just want to collapse."<br>**User reaction:** resistance / helplessness *(defensive / withdrawal)* |
| **Paradigm B:** EFT-CoT ❤️ **(Embodied Empathy)**<br>Agent: "That heaviness... it sounds like your body is exhausted. Maybe it's okay to be a pile of mud for a moment. You don't have to stand up right now. This numbness is actually your body trying to shield you from pain. Let's just gently breathe into that softness." | 🥹 User: "Finally... I don't have to fight anymore. He understands that my 'collapsing' isn't a failure, but a need for rest."<br>**User reaction:** relaxation / openness *(acceptance / trust)* |

**Fig. 1.** A comparative example of two generation strategies (CBT-style vs EFT-style) in a high-arousal context.

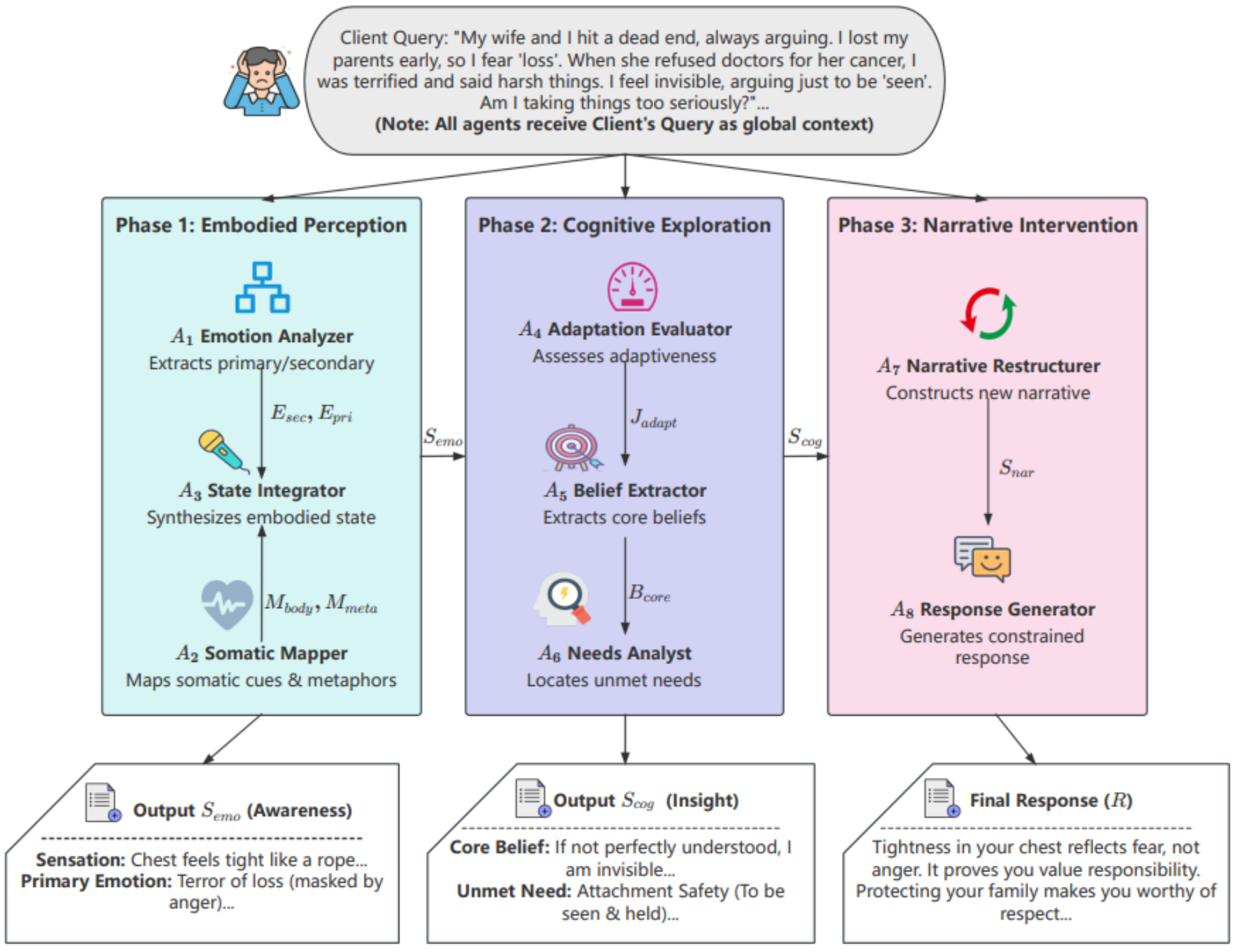

**Fig. 2.** The overall process flow of EFT-CoT. The output boxes for each stage are schematic summaries, intended to illustrate the form of intermediate states

## 2. Related Work

### *2.1. Large Language Models for Psychological Counseling*

Existing research on large language models (LLMs) for psychological counseling has advanced along two primary directions. First, interaction paradigms are evolving from single-turn question answering toward multi-turn dialogues. Second, resource construction is shifting from general-purpose corpora to high-quality datasets that are better aligned with counseling contexts, with controllable synthesis increasingly adopted to improve both scale and quality.

To address the limitation that publicly available resources are largely single-turn question–answer pairs and thus cannot capture long-horizon interactions, SMILE (Qiu et al., 2024) extends single-turn long-text question answering to multi-turn psychological support dialogues. SoulChat (Chen et al., 2023) further strengthens a model's capabilities in active listening, emotional comforting, and supportive expression by leveraging a large-scale multi-turn empathic corpus. Building on these efforts, subsequent work aims to develop more professional dialogue behaviors that are more tightly grounded in counseling scenarios. For example, ChatCounselor (Liu et al., 2023) and MentalChat16K (J. Xu et al., 2025) construct high-quality datasets from real or simulated counseling sessions, shifting the research focus from generic emotional comfort toward the enhancement of more professional counseling competencies.

Due to the scarcity of real counseling data and stringent privacy constraints, some studies have explored data construction strategies based on synthesis and reconstruction. CpsyCoun (Zhang et al., 2024) draws on publicly available counseling case reports and improves dialogue completeness and professional expression through a two-stage reconstruction pipeline. PsyDial (Qiu & Lan, 2025) desensitizes and masks real counseling records and then uses a model to reconstruct conversational details, aiming to preserve the complex interaction logic of authentic sessions as much as possible while reducing privacy risks. Overall, these studies have made progress in multi-turn interaction modeling and dataset construction; however, most methods still primarily emphasize learning dialogue formats and surface-level linguistic styles. To further improve the professionalism and structure of counseling responses, recent work has begun to incorporate professional methods from psychotherapy schools into the model's reasoning process.

### *2.2. Integration of Psychotherapeutic Approaches into LLMs*

To improve the professionalism and controllability of counseling-oriented generation, recent research has explored integrating psychotherapeutic theories more deeply into model reasoning, moving from generic empathic replies toward structured and actionable intervention steps. Owing to its clear procedural pathway and strong operability, Cognitive Behavioral Therapy (CBT) has become a mainstream entry point. For example, CBT-LLM (Na, 2024) introduces CBT strategies into MHQA via prompting to generate structured supportive responses; however, the overall approach remains relatively templated, with limited flexibility and precision in practice. Subsequent work has further focused on stage-wise and end-to-end modeling of the intervention process. HealMe (Xiao et al., 2024) decomposes CBT into explicit stages and guides users through cognitive restructuring in a three-turn dialogue format. CACTUS (Lee, Mac Kim, et al., 2024) constructs multi-turn CBT counseling dialogue data and improves process coherence through conversation planning. In addition, CRISP (Zhou et al., 2025) proposes mechanism-level refinements to cognitive restructuring, aiming to expand the interactive depth of effective intervention dialogues through a multi-channel cyclic design.

To address the limitations of single-model solutions for complex psychological tasks, multi-agent collaboration has become an active research direction. By assigning distinct roles and coordinating task execution, multi-agent systems can substantially improve the quality of counseling responses. Lee, Kang, et al. (2024) reported that, under matched test conditions, multi-agent systems outperform single-agent baselines on dimensions such as content quality, logical coherence, and user acceptance. Building on this line, MADP (Chen & Liu, 2025) enhances empathic emotional support by simulating interactions among multiple psychological elements in a CBT setting. AutoCBT (A. Xu et al., 2025) introduces dynamic routing and memory mechanisms to strengthen agents' autonomous decision-making for CBT intervention strategies. CoCoA (Lee, Kang, et al., 2024) similarly incorporates memory and applies CBT techniques to identify cognitive distortions in user statements. Overall, these methods improve intervention structure and expression quality through staged procedures and multi-role collaboration.

However, existing work still exhibits notable limitations. Most CBT-based approaches prioritize cognitive correction, using cognitive restructuring to revise users' negative thoughts, which can weaken the central role of emotional experience in psychological change. Particularly in contexts dominated by high arousal or deep emotional pain, relying solely on cognitive restructuring may be insufficient to address the help-seeker's present-moment experience. Many methods implicitly treat emotions as disturbances to be rapidly alleviated, lacking mechanisms for deeper emotional processing and transformation. This modeling bias—favoring cognition over experience—makes it difficult to reach the deeper roots of the help-seeker's suffering and to establish a more profound emotional connection. Therefore, although prior work has advanced emotional support and strategy control, process-based modeling of EFT's core emotional processing mechanisms to constrain generation remains relatively underexplored, which is the focus of this paper.

## 3. Methodology

This section presents the EFT-CoT computational framework. We first formalize EFT's core mechanisms as a phased state-transition reasoning process, then introduce the multi-agent collaborative architecture, and finally describe each agent's definition and algorithmic implementation.

### *3.1. Theoretical Foundation and Problem Formalization*

#### *3.1.1. Theoretical Foundation: Emotion-Focused Therapy (EFT)*

To support deep emotional processing in MHQA, we adopt Emotion-Focused Therapy (EFT) as the guiding theoretical framework. EFT emphasizes that emotions drive psychological change and advocates a "bottom-up" intervention path of "awareness–regulation–transformation." In our framework, these theoretical stages correspond to Phase 1–3 (Embodied Perception, Cognitive Exploration, and Narrative Intervention) and are operationalized as structured intermediate reasoning states $S_{emo}$, $S_{cog}$, and $S_{nar}$, respectively:

- **Awareness**: The core task of this stage is to help the user recognize and become aware of their present emotional experience. The model first identifies the secondary emotions (e.g., frustration) presented in the text and further uncovers the primary emotions (e.g., helplessness) behind them. It helps articulate the user's vague bodily discomforts (such as tension, heaviness, etc.) into expressible emotional representations, thereby enabling the user to more clearly recognize the source and meaning of their emotions. This stage is operationalized as the Embodied Perception state $S_{emo}$, providing the affective grounding for subsequent reasoning.
- **Regulation**: The task of this stage is to assess the emotional adaptiveness within the current context and, if necessary, help the user restore psychological resilience. The model identifies potentially maladaptive emotional responses (e.g., excessive anxiety or shame) and infers the core beliefs and unmet psychological needs behind them. Based on these inferences, the model forms structured insights that support subsequent response generation, helping the user gain distance from, and understanding of, their state of "emotional overwhelm." This stage is operationalized as the Cognitive Exploration state $S_{cog}$, providing cognitive support for emotional transformation.

- **Transformation**: The task of this stage is to facilitate emotional processing and meaning reconstruction based on prior insights. The model integrates core needs and key experiences into a new narrative framework that is more realistic and self-compassionate, introducing more adaptive emotional orientations (such as self-compassion, understanding, and hope) to loosen the original negative feedback loop (the self-reinforcing cycle of emotion—thought—behavior). This stage is operationalized as the Narrative Intervention state $S_{nar}$, supporting the final response generation.

At the computational implementation level, we operationalize the above stage-wise mechanism as a three-phase reasoning workflow—Embodied Perception ($S_{emo}$), Cognitive Exploration ($S_{cog}$), and Narrative Intervention ($S_{nar}$).

### *3.1.2. Task Definition and Mathematical Modeling*

Based on the above theoretical mappings, we formalize the MHQA task as a hierarchical state transition process. Unlike traditional end-to-end generative models $P(R \mid I)$, this study introduces three continuous intermediate state variables: $S_{emo}$, $S_{cog}$, and $S_{nar}$. The system's ultimate goal is to maximize the following joint probability:

$$R^* = \arg\max_R \sum_S P(R|S_{nar}, I) P\left(S_{nar}\middle|S_{cog}\right) P\left(S_{cog}\middle|S_{emo}\right) P(S_{emo}|I)$$

where $S = (S_{emo}, S_{cog}, S_{nar})$, $I$ is the initial input from the help-seeker, and $R$ is the final response. The $\sum_S$ term indicates marginalization over the combinations of stage states, thus incorporating intermediate reasoning uncertainties into the objective function. This formula characterizes the phased dependency structure of the reasoning process ($S_{emo} \rightarrow S_{cog} \rightarrow S_{nar}$), constraining the generation to follow the phased intervention logic of EFT. The process is shown in Algorithm 1.

| **Algorithm 1:** EFT-CoT Inference Process |
|---|
| **Input:** User Text $I$<br>**Output:** Therapeutic Response $R$ |
| 1: **Phase 1:** Embodied Perception (Parallel Processing)<br>2: parallel do<br>3: $\left(E_{sec}, E_{pri}\right) \leftarrow A_1(I)$ // primary/secondary emotions |

4: $\left(M_{body}, M_{meta}\right) \leftarrow A_2(I)$ // somatic cues & metaphors
5: end parallel
6: $S_{emo} \leftarrow A_3\left(I, E_{sec}, E_{pri}, M_{body}, M_{meta}\right)$ // awareness state
7: **Phase 2:** Cognitive Exploration (Sequential)
8: $J_{adapt} \leftarrow A_4(I, S_{emo})$// adaptiveness assessment
9: $B_{core} \leftarrow A_5\left(I, S_{emo}, J_{adapt}\right)$// core beliefs (evidence-grounded)
10: $N_{psy} \leftarrow A_6(I, S_{emo}, B_{core})$ // unmet needs (SDT)
11: $S_{cog} \leftarrow \{J_{adapt}, B_{core}, N_{psy}\}$ // insight state
12: **Phase 3:** Narrative Intervention & Response Generation
13: $S_{nar} \leftarrow A_7\left(I, B_{core}, N_{psy}\right)$ // narrative reframing
14: $R \leftarrow A_8\left(I, S_{emo}, S_{cog}, S_{nar}\right)$ // constrained generation (3 anchors)
15: return R

### *3.2. EFT-CoT Multi-Agent Collaborative Framework*

To solve the above probabilistic model, this study proposes the EFT-CoT framework, as shown in Fig. 2. The framework employs multi-agent collaboration to balance reasoning complexity with deployment usability. Specifically, the framework includes a set of dedicated agents: $\mathcal{A} = \{A_1, A_2, \dots, A_8\}$, where each agent is responsible for a specific task in the reasoning process. These agents are assigned distinct reasoning roles and jointly produce structured intermediate states and the final response. Appendix A provides a complete example of stage-wise state transitions for a counseling case under the EFT-CoT framework.

This section will describe the reasoning process within the framework, with a focus on the tasks, formal definitions, and key constraints of each phase:

- **Step 1: Embodied Perception** ($S_{emo}$): Agents $A_1 \sim A_3$ are responsible for mapping the unstructured input text $I$ into a perception state containing emotional layers and bodily metaphors.
- **Step 2: Cognitive Exploration** ($S_{cog}$): Based on the perception state,agents $A_4 \sim A_6$ infer the help-seeker's core beliefs and psychological needs, thereby generating the cognitive state.
- **Step 3: Narrative Intervention** ($S_{nar}$): Agents $A_7 \sim A_8$ transform the cognitive insights into a supportive new narrative and ultimately generate the response $R$.

To mitigate information loss in multi-hop reasoning, we treat the original input $I$ as a global context variable and explicitly inject it as shared context into each agent's input.

### *3.3. Phase 1: Embodied Perception*

This phase aims to establish an empathy baseline by extracting emotion labels and bodily cues to help the user become aware of their emotional experience.

**(1) Emotional Hierarchy Identification** ($A_1$). From an EFT perspective, agent $A_1$ decomposes the user input $I$ into a set of secondary emotions $E_{sec}$ and a set of primary emotions $E_{pri}$:

$$(E_{sec}, E_{pri}) = A_1(I)$$

**(2) Somatic Awareness Mapping** ($A_2$). Agent $A_2$ performs bodily cue mapping, generating descriptions of bodily sensations $M_{body}$(e.g., "tightness in the chest") and embodied metaphors $M_{meta}$(e.g., "like drowning"):

$$(M_{body}, M_{meta}) = A_2(I)$$

**(3) Affective State Integration** ($A_3$). Agent $A_3$ integrates the above emotion and bodily cues into a coherent first-person experiential state $S_{emo}$, providing an experiential grounding for subsequent reasoning:

$$S_{emo} = A_3(I, E_{sec}, E_{pri}, M_{body}, M_{meta})$$

### *3.4. Phase 2: Cognitive Exploration*

This phase aims to analyze mechanisms underlying psychological blockage through a serial reasoning chain, helping the user transition from "feeling" to "understanding."

**(1) Adaptive Evaluation** ($A_4$): Agent $A_4$ evaluates the functionality of the emotion and outputs an **adaptiveness** judgment result $J_{adapt}$:

$$J_{adapt} = A_4(I, S_{emo})$$

**(2) Core Belief Extraction** ($A_5$): To mitigate hallucinations, agent $A_5$ uses evidence-based inference strategies to abstract core beliefs $B_{core}$(e.g., "I am incompetent") from the narrative that drive negative emotions:

$$B_{core} = A_5(I, S_{emo}, J_{adapt})$$

**(3) Needs Analysis** ($A_6$): Based on Self-Determination Theory (SDT), agent $A_6$ identifies unmet deep psychological needs $N_{psy}$(e.g., lack of belonging):

$$N_{psy} = A_6(I, S_{emo}, B_{core})$$

At this point, we obtain the complete cognitive state $S_{cog} = \{J_{adapt}, B_{core}, N_{psy}\}$.

### *3.5. Phase 3: Narrative Intervention*

This phase transforms structured reasoning into a natural language intervention $R$ to address the user's emotional issues.

**(1) Narrative Restructuring** ($A_7$): Agent $A_7$ compares the old narrative (i.e., the core belief $B_{core}$) with the unmet needs $N_{psy}$ and constructs a new narrative framework $S_{nar}$ that balances realism and resilience, achieving meaning reconstruction. It reframes negative experience into more adaptive psychological activity:

$$S_{nar} = A_7(I, B_{core}, N_{psy})$$

**(2) Constrained Response Generation** ($A_8$): Agent $A_8$ aggregates the global context $I$ with the reasoning states from all previous stages ($S_{emo}, S_{cog}, S_{nar}$) to generate the final response $R$. To ensure the professionalism and logical closure of the response, the generation process, based on full information, is explicitly constrained by three "anchors": empathic metaphors ($M_{meta} \in S_{emo}$), deep needs ($N_{psy} \in S_{cog}$), and the new narrative framework ($S_{nar}$):

$$R = A_8(I, S_{emo}, S_{cog}, S_{nar})$$

This mechanism ensures that the final response is not only semantically coherent but also covers the entire emotional processing logic, from perception to transformation.

## 4. Dataset Construction

### *4.1. Dataset and Preprocessing*

This study adopts the construction paradigm of the PsyQA dataset, and through systematic updates and expansions, builds a new Chinese psychological question-answering corpus covering nine major psychological topics. As an authoritative psychological counseling question-answering dataset, PsyQA integrates 22,000 historical Q&A pairs from the psychological question-answering section of the well-known Chinese psychological service platform "Yixinli." However, its data was outdated, and the quality was initially limited. After updates to the source platform's Q&A section (such as adding a tipping feature), the overall data quality improved. Therefore, this study re-collected 67,778 high-quality Q&A data points, including user questions, counselor responses, and basic metadata (such as topic tags). The content covers nine major psychological topics, including interpersonal relationships and personal growth, and the dataset is representative and diverse.

Given that the source platform has established strict content review and privacy protection mechanisms, the original data has been stripped of explicit sensitive information before being made public. Additionally, to further ensure the safety of the training process, we implemented secondary filtering using the built-in safety defense mechanisms of the base models. This includes the removal of any extreme, high-risk samples (e.g., suicide interventions) that would trigger the model's rejection. Ultimately, a high-quality corpus containing 67,398 samples was constructed.

### *4.2. Process Augmentation and Label Construction*

To allow the downstream model to implicitly learn the phased intervention logic of EFT, we adopt the EFT-CoT-driven process augmentation paradigm. Specifically, we input the 67,398 real help-seeking texts into the EFT-CoT framework to construct high-quality process-aligned training data.

#### *4.2.1. Topic-Adaptive Heterogeneous Base Strategy*

This study uses a dynamic routing strategy based on topic labels, assigning different counseling tasks to different base models to improve synthesis quality and enhance output consistency (see Table 1). To control generation randomness, we set the temperature parameter to 0.7 during the generation phase, with all other generation parameters following the official default settings of each model interface.

**Table 1**

Topic-Adaptive Heterogeneous Teacher Routing Strategy.

| Topic Category (Label) | Model | Routing Motivation (Brief Description) |
|---|---|---|
| Growth, Love, Career | Doubao-1.5-pro | Closer to youth expressions and online contexts |
| Marriage, Family, Emotion | Qwen-Max | Suitable for Chinese emotional contexts and long-text responses |
| Interpersonal, Behavior | DeepSeek-Chat | Focused on relationship structure and reasoning consistency |
| Therapy | GPT-4o | Used to reduce the risk of inappropriate advice and improve consistency |

#### *4.2.2. Internalization of Implicit Reasoning Logic*

Using the above heterogeneous expert set, we fully execute the eight-stage reasoning chain shown in Fig. 2, expanding the original input $I$ into a tuple containing the complete phased reasoning process ($I \rightarrow CoT_{A1 \ldots A7} \rightarrow R$). However, when constructing the final

EFT-Instruct corpus, we strategically removed all explicit intermediate states ($CoT_{A1...A7}$) and only retained the "input—final response" pairs ($I \rightarrow R$).

It is important to note that the intermediate states generated by EFT-CoT are used for process auditing and data quality control, while EFT-LLM does not require explicit outputs. The core motivation behind this design is the "implicitization of efficiency and logic"—aiming to allow the model to directly learn the high-dimensional mapping from "problem representation" to "EFT-compliant response." In this way, the model can internalize the complex "slow thinking" (EFT state transition logic) into efficient "fast intuition" (latent parameter distributions), thereby significantly reducing token consumption and latency in the reasoning phase while ensuring high-quality professional interventions. This design is particularly suited for resource-constrained scenarios.

The goal of the EFT-CoT framework is not only to generate auditable process states but also to construct a process-aligned dataset, EFT-Instruct. To enable this intervention logic to be replicated in real dialogues in a low-cost, stable manner, we further fine-tuned EFT-Instruct to obtain EFT-LLM, internalizing the "Embodied Perception—Cognitive Exploration—Narrative Intervention" strategy into the model parameters. Correspondingly, in the experimental section (RQ3), we examine the training effectiveness of EFT-Instruct, assessing whether it improves the model's performance in professional intervention dimensions and demonstrates transferability when compared to strong baselines.

## 5. Experiments

In this section, we conduct a comprehensive evaluation of the EFT-CoT multi-agent system and the fine-tuned model EFT-LLM. We focus on answering three research questions:

- **RQ1 (Framework Effectiveness)**: Does EFT-CoT demonstrate significant advantages in EFT-specific dimensions compared to the single-agent CoT baseline (Single-agent CoT)?
- **RQ2 (Ablation of Mechanisms)**: Do the key modules in the EFT-CoT framework each yield their intended effects?
- **RQ3 (Model Performance)**: Is EFT-LLM superior to existing strong baseline models?

To validate the above hypotheses, we expect to observe three points:

- The gains in RQ1 should mainly be concentrated in EFT-related dimensions such as embodied perception and narrative restructuring;

- RQ2's ablation should lead to directional degradation consistent with the function of the modules;
- The advantages in RQ3 should primarily manifest in empathy depth and narrative restructuring, maintaining reliable inter-rater consistency across quality intervals.

### *5.1. Implementation Details and Evaluation Protocol*

#### *5.1.1. Implementation Details*

- EFT-LLM is based on Qwen-2.5-7B-Instruct (Yang et al., 2025), fine-tuned with LoRA (Rank=16, Alpha=32) using LLaMA Factory(Hu et al., 2022; Zheng et al., 2024) and BF16 mixed precision. Other hyperparameters are listed in Appendix B. Training uses the "implicit chain of thought" strategy, with only the final response $R$ as the supervisory signal.
- Inference Configuration: To strictly control variables in RQ1 and RQ2 and ensure performance differences arise from architectural design rather than model capacity, the above inference experiments uniformly use DeepSeek-chat (Temperature=0.7). RQ3 comparison experiments also use Temperature=0.7.

#### *5.1.2. Data Preparation*

EFT-Instruct is randomly split into training/test sets in a 9:1 ratio. During evaluation, two complementary sets are constructed:

- **Full Test Set** (N=6,740): Used to calculate general automation metrics.
- **Core Evaluation Set** (N=180): Stratified sampling based on nine major topics, used for evaluating psychological professional dimensions.

#### *5.1.3 General Evaluation Protocol*

(1) **Automated Metrics (Full Test Set)**: BLEU (Papineni et al., 2002), ROUGE (Lin, 2004), BERTScore (Zhang et al., 2020), METEOR (Banerjee & Lavie, 2005), Distinct(Li et al., 2016) (as supplementary references for generation quality).

(2) **Professional Dimension Evaluation (Core Evaluation Set)**: Given that high-performance model evaluations often exhibit high score clustering and low variance, traditional consistency metrics based on correlation may fail. Therefore, we use AAD (Average Absolute Deviation) and Adjacent Agreement Rate (score difference ≤ 1) to quantify consensus and stability across different performance intervals. In addition, to eliminate position bias and ensure the objectivity and rigor of the scores, we add explicit debiasing constraints to the judge prompts and anonymize the model identity of the candidate responses. Subjective evaluation is independently scored by a multi-model panel consisting of GPT-4o, Claude-sonnet-4-5-20250929 (hereinafter referred to as Claude-4.5), and DeepSeek-chat. Similarly, previous psychological counseling data construction tasks have used GPT-4 as a judge to compare and evaluate the generated dialogues (Zhang et al., 2024). On the other hand, recent systematic research on "empathy communication evaluation" indicates that, under the appropriate framework and benchmarks, LLMs can achieve expert-level consistency in scoring subtle empathic differences, often outperforming crowd-sourced annotations, thus serving as a scalable evaluation tool for emotionally sensitive tasks (Kumar et al., 2026). Based on this, statistical significance is assessed using one-sided Wilcoxon signed-rank paired tests (Ours > Baseline/Variant).

### *5.2. Experiment 1: Evaluation of Framework Effectiveness (RQ1)*

This section aims to verify whether the proposed EFT-CoT framework achieves performance gains over Single-agent CoT on professional dimensions aligned with Emotion-Focused Therapy (EFT).

#### *5.2.1. Experimental Setup*

Using the core evaluation set ($N = 180$), we conducted an anonymized, paired comparative experiment. The compared baselines are as follows:

- **Single-agent CoT (Baseline):** A prompt-engineering baseline designed to contrast two implementation pathways: a *single-pass CoT prompt* versus *multi-agent procedural reasoning*. This approach adopts a single-agent architecture and uses a system prompt incorporating EFT principles (see Appendix C.1) to guide the model, within a single generation, to follow a generic three-step chain of thought: emotion identification → needs analysis → empathic intervention.
- **EFT-CoT (Ours):** The proposed collaborative framework consisting of eight agents.

#### *5.2.2. Evaluation Metrics*

Because automated metrics struggle to capture the nuance of psychological intervention, this experiment primarily adopts a five-dimensional professional evaluation system grounded in EFT theory: somatic awareness, emotional hierarchy, adaptive insight, needs analysis, and narrative restructuring capacity. The theoretical origins of each dimension and the detailed scoring criteria are provided in Appendix C.1.

#### *5.2.3. Results Analysis*

First, the consistency check indicates that the global AAD is only 0.307 (adjacent agreement rate 99.2%), suggesting high inter-rater consistency in scoring responses of varying quality and providing a reliable basis for subsequent comparisons.

As shown in Table 2, EFT-CoT demonstrates significant advantages across all evaluated dimensions (mean 4.91 vs. 3.44, $p < 0.001$), strongly supporting the effectiveness of a multi-agent architecture for counseling-oriented tasks. In particular, the most substantial qualitative shift occurs in the perceptual dimension: somatic awareness increases by +2.34 points (4.80 vs. 2.46). This large gap is primarily attributable to the introduction of embodied perception agents ($A_1$–$A_2$), which partially mitigates the limitations of general-purpose LLMs in expressing bodily cues in purely textual contexts, thereby biasing the responses toward context-congruent embodied metaphors.

Building on this foundation, the model gains further benefits from the explicit decoupling of cognitive processes: the near-ceiling performance in adaptive insight (4.99) and needs analysis (4.99) confirms the necessity of this strategy. Unlike Single-agent CoT, which tends to fall into the trap of premature advice-giving, EFT-CoT enforces the logical constraint of "assessment before intervention," enabling more precise identification of the help-seeker's core beliefs. Notably, the advantage in narrative restructuring capacity (4.99) does not emerge in isolation; rather, it reflects a positive downstream propagation of high-quality reasoning from earlier modules. With prior deep empathy and belief elicitation in place, the model's suggestions are no longer rigid or didactic; instead, they more naturally guide the help-seeker toward reconstructing meaning—shifting from a "victim" frame to a "survivor" frame.

**Table 2**

Comparison of EFT-CoT and Single-agent CoT on core counseling-related dimensions ($N = 180$).

| Dimension | EFT-CoT | Single-agent CoT |
|---|---|---|
| Somatic Awareness | 4.80 | 2.46 |
| Emotional Hierarchy | 4.77 | 3.70 |
| Adaptive Insight | 4.99 | 3.75 |
| Needs Analysis | 4.99 | 3.84 |
| Narrative Restructuring | 4.99 | 3.43 |
| Average | 4.91 | 3.44 |

### *5.5. Experiment 2: Mechanism Ablation (RQ2)*

#### *5.5.1. Experimental Setup*

We conduct a mechanism ablation study via functional blocking to examine the computational irreducibility of the three-stage process in EFT-CoT—Embodied Perception → Cognitive Exploration → Narrative Intervention. It should be emphasized that this experiment is not intended to argue that multi-agent architectures are inherently superior. Rather, we treat each agent as an executable unit that instantiates a specific EFT mechanism. Accordingly, physically removing an agent is essentially a way to test whether the corresponding theoretical step constitutes a necessary prerequisite for generating high-quality psychological interventions.

- **w/o Somatic (blocking the embodied mechanism):** We remove the somatic mapping agent $A_2$, which blocks the mapping from textual cues to embodied cues/metaphors: $(M_{body}, M_{meta}) = A_2(I)$. This variant tests whether deep empathy must rely on physiologically grounded, concretized perception, or whether emotional processing can be achieved using textual semantics alone.
- **w/o Cognition (blocking the cognitive mechanism):** We remove the deep cognitive inference modules, including core-belief extraction ($A_5$) and needs analysis ($A_6$). While this variant retains surface-level emotion recognition, it cannot explicitly infer the core belief $B_{core}$ or the psychological need $N_{psy}$. This setting simulates a form of "shallow counseling" that lacks deeper psychodynamic analysis, and is designed to test whether deep cognitive insight is necessary for producing effective guidance.
- **w/o Restructuring (blocking the narrative mechanism):** We remove the narrative restructuring agent $A_7$. This variant preserves the state outputs from earlier stages $(S_{emo}, S_{cog})$, but blocks the critical step that converts cognitive exploration outcomes into a supportive alternative narrative, thereby testing the role of narrative intervention in closing the intervention loop.

#### *5.5.2. Results Analysis*

First, inter-rater consistency remains high, with a global AAD of 0.344 (adjacent agreement rate 99.1%). This indicates strong consensus when evaluating different variants, ensuring that subsequent differences primarily reflect genuine degradations caused by functional blocking rather than rater noise.

As shown in Table 3, the full EFT-CoT achieves the best performance across all dimensions (overall mean 4.82, $p < 0.001$). This result provides initial support for RQ2, suggesting that EFT-aligned intervention quality depends on the complete synergy of the embodied perception–cognitive exploration–narrative intervention pipeline, and that removing any stage leads to a significant performance decline.

**(1) w/o Somatic.** Removing $A_2$ is equivalent to cutting off the computational pathway from textual cues → embodied representations. The largest drop is observed in Somatic Awareness (4.58 → 3.52, Δ = −1.06), consistent with the targeted degradation expected from the absence of explicit embodied mapping. Without this mechanism, the model has greater difficulty generating embodied cues that align with lived experience, and empathic responses are more likely to remain at an abstract semantic level. Notably, the loss of bodily cues also produces an interpretable cascading effect: Emotional Hierarchy decreases to 4.05, indicating that embodied perception influences not only *how* empathy is expressed but also *where* empathic attunement is anchored.

**(2) w/o Cognition.** Removing $A_5$ and $A_6$ blocks explicit inference of the core belief $B_{core}$ and psychological need $N_{psy}$. This variant exhibits the most pronounced overall degradation (mean 4.03, tied for the lowest with w/o Somatic), indicating that cognitive exploration is a decisive component of intervention quality. The data reveal a key mechanistic pattern: even though the narrative module is retained, Narrative Restructuring still drops to 4.06, precisely matching the expected directional degradation under cognitive blocking. This finding suggests that intervention quality is determined less by how the response is phrased (rhetorical style) than by what is said (the insight anchor). When belief- and need-level insights are missing, narrative expression tends to lose target specificity and degenerates into superficial reassurance or generic advice.

**(3) w/o Restructuring.** This variant retains earlier perceptual and insight states but removes the explicit step $A_7$ that translates insights into a supportive alternative narrative. As a result, Narrative Restructuring decreases significantly (4.92 → 4.33, $p < 0.001$), consistent with the targeted degradation expected from removing narrative intervention. This confirms the critical role of the narrative agent in bridging the gap between *intervention analysis* and *user receptivity*: accurate understanding alone does not guarantee effective supportive expression. Meaning reconstruction must be realized through $A_7$; otherwise, the model cannot reliably guide users toward deeper psychological change.

Overall, the three functional blockings exhibit targeted degradation patterns that align with the hypothesized roles of the corresponding modules, thereby validating the proposed mechanism: embodied perception provides experiential grounding and supplies somatic cues and embodied metaphors; cognitive exploration localizes core beliefs and psychological needs; and narrative intervention translates cognitive states into a receivable alternative narrative. Together, these stages form an irreducible mechanistic chain that operates synergistically under structured state-flow constraints, supporting process alignment and the final response quality of EFT-CoT.

**Table 3**

Results of the mechanism ablation study: validating the irreducibility of functional modules in the EFT-CoT framework ($N = 180$). The best results are highlighted in bold; "↓" marks the dimension(s) exhibiting the largest (or tied-largest) performance degradation for each variant; Som/Cog/Res denote somatic mapping, cognitive exploration, and narrative restructuring, respectively, and "–" indicates module removal (ablation).

| Metric | EFT-CoT | w/o Som ($A_2$) | w/o Cog ($A_5$, $A_6$) | w/o Res ($A_7$) |
|---|---|---|---|---|
| Somatic Awareness | 4.58 | 3.52 ↓ | 3.97 | 4.12 |
| Emotional Hierarchy | 4.73 | 4.05 | 3.94 | 4.17 |
| Adaptive Insight | 4.90 | 4.18 | 4.09 ↓ | 4.34 |
| Needs Analysis | 4.96 | 4.28 | 4.09 ↓ | 4.41 |
| Narrative Restructuring | 4.92 | 4.13 | 4.06 | 4.33 ↓ |
| Average | 4.82 | 4.03 | 4.03 | 4.27 |

### *5.3. Experiment 3: Model Performance Evaluation (RQ3)*

This section evaluates the overall performance of EFT-LLM in practical counseling scenarios through a horizontal comparison with representative baselines.

#### *5.3.1. Experimental Setup*

To ensure a comprehensive evaluation, we selected four representative baselines:

- **Qwen-2.5-7B-Instruct (hereafter, Qwen):** The base model used to quantify domain gains brought by instruction tuning.
- **GPT-4o + Standard Prompting:** A strong baseline representing a general-purpose LLM under standard prompting conditions. We adopt an open-source, commonly used counseling prompt (e.g., "I want you to act as a psychologist…") to elicit responses.

- **CBT-Prompting:** To compare against a strong, representative baseline in MHQA, we introduce a prompting baseline based on the Cognitive Behavioral Therapy (CBT) intervention paradigm. Specifically, we reuse the prompting framework proposed in CBT-LLM to obtain a reproducible CBT-style reasoning process. However, because CBT-LLM publicly releases model weights but not training data, and because it was trained on Baichuan-7B, directly using the original model would confound the comparison with differences in base-model capability, thereby undermining an objective assessment of the intervention paradigm itself. Therefore, we implement the CBT-LLM prompting framework on Qwen for inference, enabling a more controlled comparison between the CBT prompting paradigm and our method under a consistent base model.
- **PsyQA:** Real human-written responses from the "YiXinLi" community, used as a real-world reference anchor.

#### *5.3.2. Evaluation Metrics*

In contrast to the micro-level analysis of EFT-specific mechanisms in RQ1, this experiment aims to evaluate the model's end-to-end performance in real-world psychological support scenarios from a macro-level, general counseling perspective. To this end, we adopt four comprehensive evaluation dimensions widely recognized in existing mental health LLM research: Relevance, Empathic Depth, Helpfulness, and Structural Professionalism.

Specifically, *Relevance* evaluates how closely the response aligns with the user's core distress and contextual details, ensuring a hallucination-free output. *Empathic Depth* measures the model's capacity to understand and accept deep-seated emotions and to establish a strong working alliance. *Helpfulness* assesses the practical utility of the response in alleviating emotional distress, providing effective support, or offering novel perspectives. Finally, *Structural Professionalism* examines whether the response adheres to a coherent supportive rhythm and evidence-based workflows (e.g., avoiding "premature advice-giving" in favor of "progressive exploration and intervention"). These four metrics are progressively linked, jointly constituting a comprehensive evaluation loop that ranges from foundational dialogue quality to advanced professional support (detailed scoring criteria are provided in Appendix C.2).

#### *5.3.3. Results and Analysis*

**(1) General Generation Quality (Automatic Metrics)**

EFT-LLM outperforms the base model on all automatic metrics. Notably, these metrics mainly reflect semantic overlap and surface-level generation characteristics, and cannot fully capture the effectiveness of psychological interventions. We therefore treat them as supplementary evidence, while relying primarily on the EFT-grounded professional evaluation for our main conclusions. The automatic evaluation results on the test set are summarized in Table 4.

The results show substantial improvements in BERTScore (88.35) and ROUGE-L (36.80), indicating that the model not only acquires an EFT-consistent linguistic style but also aligns more accurately with deeper semantic logic. Moreover, the high Distinct-3 (93.24) suggests that, after LoRA fine-tuning, the model maintains strong response diversity without exhibiting the common issue of mode collapse.

**Table 4**

Automated evaluation results on the test set. Except for BERTScore, other metrics are evaluated by Chinese character-level tokenization. For BERTScore, we use the BAAI/bge-m3 model to get text embedding.

| | METEOR↑ | B-1↑ | B-2↑ | B-3↑ | R-L↑ | D-1↑ | D-2↑ | D-3↑ | BERTScore↑ |
|---|---|---|---|---|---|---|---|---|---|
| Qwen | 34.02 | 38.43 | 22.10 | 12.69 | 17.76 | 33.46 | 70.59 | 86.13 | 78.96 |
| EFT-LLM | 47.86 | 61.70 | 45.72 | 35.29 | 36.80 | 42.70 | 82.45 | 93.24 | 88.35 |

**(2) Counseling Intervention Quality (Professional Metrics)**

First, the consistency check indicates that the judges' scoring criteria remained highly stable even after introducing a lower-performing base model (Qwen) and unstructured human responses (PsyQA). The global AAD is 0.589, and 92.8% of score differences are within 1 point, supporting the reliability of LLM-based judging across a wide quality range. On this basis, EFT-LLM achieves the best overall performance (mean score 4.36), significantly outperforming all baselines ($p < 0.001$). Detailed results are reported in Table 5. The comparison further reveals substantive differences across intervention pathways:

**(i) EFT-LLM vs. general-purpose LLMs**

Both GPT-4o and Qwen lag substantially in empathic depth, suggesting that general-purpose models without explicit theoretical alignment tend to produce mechanistic advice and remain confined to an "intellectualized" response style, thereby failing to provide deeper emotional support.

**(ii) EFT-LLM vs. CBT-Prompting**

Given the same inputs, CBT-Prompting outputs are more concentrated on cognitively oriented, structured analysis (e.g., problem identification → alternative thoughts → behavioral recommendations), while scoring markedly lower in empathic depth ($\Delta \approx 1.38$). This pattern is consistent with CBT's top-down paradigm: when a prompt explicitly requires multi-step cognitive tasks, the model prioritizes completing an "analysis–correction" discourse structure, which compresses the expressive space for fine-grained attunement to the user's immediate experience. In contrast, EFT-LLM internalizes a "experience-first, transformation-second" strategy through fine-tuning, leading to more stable performance in empathic depth and narrative restructuring.

**(iii) EFT-LLM vs. PsyQA**

The relatively lower score of PsyQA (3.12) reflects the high heterogeneity of real community responses in style, length, and strategy (e.g., a stronger tendency toward direct advice), which results in larger variance under our rubric that emphasizes a structured intervention process. This observation, in turn, highlights the value of process-aligned data for output normalization: the EFT-Instruct dataset encodes an "emotion–belief–transformation" reasoning chain that is substantially more normative and intervention-consistent than unfiltered Internet corpora.

Overall, the performance advantage of EFT-LLM is closely tied to EFT-CoT-driven, process-enhanced data construction. By injecting the intervention mechanism of Embodied Perception–Cognitive Exploration–Narrative Intervention into the training signal, the model becomes more robust on professional metrics while producing responses with greater emotional warmth.

**Table 5**

Performance comparison on professional counseling metrics (Core Evaluation Set, N=180). *Note:* Rel.: Relevance; Emp.: Empathetic Depth; Help.: Helpfulness; Struct.: Structural Professionalism. Comparison baselines include CBT-LLM, GPT-4o, Qwen-2.5-7B (Base), and human responses (PsyQA). The best scores are in bold.

| Model | Rel. | Emp. | Help. | Struct. | Avg. |
|---|---|---|---|---|---|
| EFT-LLM | **4.95** | **4.99** | **4.63** | **4.92** | **4.87** |
| Qwen | 3.10 | 2.21 | 2.93 | 2.71 | 2.74 |

| GPT-4o | 3.84 | 3.28 | 3.62 | 3.67 | 3.60 |
|---|---|---|---|---|---|
| CBT-LLM | 4.25 | 3.82 | 4.22 | 4.25 | 4.14 |
| PsyQA | 3.05 | 2.72 | 2.76 | 2.44 | 2.74 |

## 6. Process Validity and Interpretability Analysis

Unlike traditional end-to-end generative models that are evaluated primarily through black-box outcome quality, this study argues that psychological support generation should additionally emphasize process validity: whether the model's reasoning trajectory follows the prescribed staged EFT pathway, and whether key intermediate reasoning states are traceable and logically justified. Based on the core evaluation set ($N = 180$), we conduct a white-box audit of the intermediate states produced by EFT-CoT along its eight-stage reasoning chain ($A_1$–$A_8$). Under the same evaluation setting as in Chapter 5 (with the same panel of judges and consistent metric definitions), we construct a hybrid audit protocol that combines binary logical judgments (0/1) with a 1–5 Likert scale (see Appendix D), and the chain outputs are re-examined by a multi-model judging panel. Table 6 summarizes the audit dimensions for stages $A_1$–$A_8$, along with pass rates/mean scores and the corresponding agreement statistics ($N = 180$).

### *6.1. Audit Protocol and Verification of Logical Completeness*

To characterize inter-judge agreement, we report two statistics for binary indicators: the logical pass rate (overall proportion of "Pass") and the absolute agreement rate (AAR) (the proportion of cases in which all three judges provide identical decisions). For 1–5 Likert-scale indicators, we report the mean score and use the average absolute deviation (AAD) to quantify disagreement, supplemented by the adjacent agreement rate (score difference ≤ 1) as an intuitive reference. Notably, when the outcome distribution of a binary task is highly skewed (e.g., almost all cases are "Pass"), κ-type agreement coefficients can be sensitive to class imbalance and yield unstable values. We therefore do not report κ and instead use AAR uniformly to describe agreement for binary judgments.

The audit results indicate high process-level correctness and strong agreement on key logical steps. Adaptive Evaluation ($A_4$) achieves a logical pass rate of 99.81% with an AAR of 99.44%; Narrative Restructuring ($A_7$) achieves a logical pass rate of 99.44% with an AAR of 98.33%. Core Belief extraction ($A_5$) is similarly robust (pass rate 98.33%, AAR 96.11%), suggesting that judges largely agree on whether belief inferences are sufficiently grounded in the input text. Together, these results support that EFT-CoT can reliably execute key steps—such as evaluating emotion function, locating core beliefs, and reframing perspective—and can provide a dependable intermediate reasoning basis for subsequent generation.

In contrast, Emotional Hierarchy Identification ($A_1$) yields a lower AAR of 83.33% than other logical validation steps, indicating remaining disagreement regarding the boundary between primary and secondary emotions. Although the overall logical pass rate reaches 89.81%, pass rates differ across judges (GPT-4o: 97.22%, DeepSeek: 84.44%, Claude: 87.78%), suggesting that this task is more susceptible to subjective thresholds when assigning borderline affective states (e.g., anxiety/worry) to hierarchy levels. This result further implies that, for borderline samples, inter-annotator consistency could be improved through more fine-grained criteria or example-based constraints in future work.

### *6.2. Integrated Gains from Embodied Empathy to Generation Quality*

At the response-quality level, quantitative results reveal an "integrated translation" effect from intermediate insights to the final response. Somatic Awareness Mapping ($A_2$) achieves a mean score of 4.17 with AAD 0.27 and an adjacent agreement rate of 98.89%, indicating that embodied metaphors are generally context-congruent with limited disagreement. Emotion Verbalization ($A_3$) achieves a mean of 4.23 with AAD 0.26, suggesting stable alignment in granularity and intensity of emotional expression. Needs Analysis ($A_6$) achieves a mean of 4.82 with AAD 0.17, demonstrating that the model can consistently anchor need states along the SDT dimensions and produce relatively specific, structured insights.

More importantly, after Constrained Response Generation ($A_8$) integrates the intermediate states from earlier stages, the final response quality further improves: empathic depth reaches 5.00 with AAD = 0.00 (i.e., zero disagreement among judges). Meanwhile, the hallucination-free rate reaches 99.63% with an AAR of 99.44%. These results indicate that EFT-CoT not only forms stable, structured intermediate insights upstream, but also translates them into highly consistent supportive expressions in the final response, thereby providing traceable and interpretable mechanistic evidence linking process-level states to outcome-level quality.

**Table 6**

Summary of the white-box audit results for intermediate states in EFT-CoT ($N$ = 180). For logical validation indicators, we report AAR (absolute agreement rate among three judges). For Likert-scale indicators, we report AAD (average absolute deviation) and Adj. agr. (adjacent agreement rate, |Δ| ≤ 1).

| Group | Agent | Audit dimension | Result | Inter-rater agreement |
|---|---|---|---|---|
| Logical validation | $A_1$ | Primary vs. secondary emotion classification correctness | Pass rate: 89.8% | AAR: 83.3% |
| | $A_4$ | Adaptive appraisal and emotion-function interpretation correctness | Pass rate: 99.8% | AAR: 99.4% |
| | $A_5$ | Evidence-grounded core-belief inference correctness | Pass rate: 98.3% | AAR: 96.8% |
| | $A_7$ | Narrative perspective shift correctness (victim → agentic self) | Pass rate: 99.4% | AAR: 98.3% |
| Quality generation | $A_2$ | Contextual fit of bodily cues and embodied metaphors | Mean (1–5): 4.17 | Adj. agr.: 98.9% (\|Δ\|≤1) |
| | $A_3$ | Accuracy, granularity, and intensity matching of emotion expression | Mean (1–5): 4.23 | AAD: 0.45 |
| | $A_6$ | Accuracy and specificity of SDT-based need identification | Mean (1–5): 4.82 | AAD: 0.17 |
| | $A_8$ | Empathic depth | Mean (1–5): 5.00 | AAD: 0.00 |
| | | Non-hallucination rate | Pass rate (0/1): 99.63% | — |

## 7. Case Study

As illustrated in Table 7, we selected a complex case involving deep attachment trauma and high-arousal emotions to demonstrate the interpretable cues in the EFT-LLM output that correspond to the EFT-CoT staged mechanisms (for the complete dialogue transcript, please refer to Appendix E). The user exhibited strong hostility and self-doubt, expressing grief related to early bereavement and an extreme fear regarding his wife's illness recurrence.

To establish a comparison, we generated a response using the CBT-Prompting baseline under identical input conditions. This baseline tends to initiate intervention via a top-down pathway of “identifying cognitive distortions → providing corrective advice.” In a high-arousal context, it leans toward a "preachy" expression style, which weakens attunement to the present experience and may even provoke user resistance. In contrast, EFT-LLM's final response exhibits interpretable cues consistent with the three-stage EFT-CoT framework (reflecting implicit strategy features rather than explicit intermediate states). Regarding the output features corresponding to Embodied Perception ($A_1$–$A_3$), the response does not merely assign emotion labels; instead, it uses empathic mirroring combined with embodied metaphors to ground the experience at a perceptible level. This sequence of "attuning to experience first, then entering interpretation" provides a stable emotional processing entry point for subsequent de-pathologization and need-focusing. During Cognitive Exploration ($A_4$–$A_6$), the response reframes “saying harsh words” as a protective reaction under a threatening situation rather than a moral defect. It further points its function toward the core threat of “avoiding another loss" and focuses the psychological need on the "longing to be responded to.” Finally, in Narrative Intervention ($A_7$–$A_8$), the response restructures the argument from being a sign of “incompatible personalities” to a protective reaction driven by the “fear of loss.” By loosening the high-pressure belief that one "must be perfectly understood," it successfully reduces the confrontational intensity of the communication.

**Table 7**

Qualitative analysis of EFT-LLM in handling complex attachment trauma cases: The implicit manifestation of EFT-CoT's core functions. Note: Different colors illustrate the implicit intervention logic within the EFT-LLM response, corresponding to key modules in the teacher framework: 1. Emotional Hierarchy Identification (Pale Cyan, corresponding to $A_3$); 2. Somatic Awareness Mapping (Light Green, corresponding to $A_2$); 3. Adaptive Assessment (Lavender, corresponding to $A_4$); 4. Core Belief Extraction (Light Yellow, corresponding to $A_5$); 5. Needs Analysis (Light Blue ,corresponding to $A_6$); 6. Narrative Restructuring (Light Red, corresponding to $A_7$).

**EFT-LLM Response Example**

**Input:** I feel like my wife and I have hit a dead end. We argue constantly, neither convincing the other, ending only in exhaustion. I know I have issues, but I can't control them. I grew up in a broken home, lost my father at 20 and my mother at 30. I am terrified of loss. Last year, my wife had surgery for thyroid cancer. Recently, she felt pain but refused my suggestion to see a doctor for conditioning. In that moment, I wasn't just angry; I was panicked. I said some harsh words.

It's the same with my work stress—when I try to talk, she doesn't respond. I feel invisible. To make her 'see' my true self, I keep arguing, but she just thinks I'm difficult. Am I too serious? Should I just hide my true self? Is wanting to be understood by my partner really too much to ask?

**EFT-LLM Response:** Reading your words "we argue and neither convinces the other," I can deeply feel that suffocating sense of being misunderstood and unseen.

Especially when thinking about your wife's pain after her recovery, this anxiety makes your emotions uncontrollable. It feels like tight ropes binding your chest, a lump of undigested food stuck in your throat, and a knot of tangled earphone wires in your stomach. This sensation is truly agonizing.

You are so afraid of failing your family. Even in the shadow of your parents' passing, you strive to protect your home. This shows you value "responsibility" far more than "willfulness." This anxiety is a reminder: you care deeply about your family, you are not "too serious" as you claim. You are so terrified of reliving the pain of loss that you'd rather argue to express your true feelings than hide them.

This emotion tries to protect you, making you feel seen; but it also blocks communication. For a long time, the belief "I must be fully understood by others" has stood like a heavy mountain, making you feel that unless she understands you perfectly, you are not worthy of being seen.

But actually, the *you* who wants her to know your true heart deserves respect. What you need is not just her understanding, but the permission to express your true feelings.

Perhaps we can try to set down the burden of "must be understood" for a moment. In that wish to "let her know my true heart," lies your true vitality. Regardless of whether you are fully understood in the end, the *you* who wants to take responsibility for himself has already grown up.

## 8. Conclusion

This study proposes EFT-CoT, a multi-agent collaborative framework for mental health question answering. EFT-CoT operationalizes the core intervention mechanisms of Emotion-Focused Therapy (EFT) into a staged reasoning workflow—embodied perception, cognitive exploration, and narrative intervention—and, based on this workflow, constructs the EFT-Instruct corpus containing approximately 67,000 instruction-tuning instances. An EFT-aligned model fine-tuned on this corpus, EFT-LLM, outperforms strong baselines under the multi-dimensional professional evaluation protocol defined in this paper. As a methodological proof of concept, our results suggest that injecting the process mechanisms of a psychotherapy paradigm into large language models via a framework–data–model pipeline is feasible, and further highlight expert-guided data synthesis as a promising route toward building high-specialization psychological support models. Importantly, this work does not constitute a clinical substitute; for deployment in real-world settings, future research must incorporate licensed therapists within an ethics- and safety-governed framework and conduct rigorous human-in-the-loop evaluation and validation.

## 9. Limitations and Future Work

Although our experiments validate the effectiveness of EFT-CoT for mental health support generation, this work remains exploratory and has several limitations that warrant further improvement. First, the clinical scope and boundaries of the proposed approach must be clearly delineated. Our system is intended to provide procedural and structured psychological support, rather than clinical diagnosis or psychotherapy. Accordingly, it does not guarantee stable clinical improvement for individual cases, nor can it fully cover the diverse counseling needs arising from variation in risk level, cultural background, and individual differences.

Second, the noise in auto-generated signals should not be overlooked. The current framework relies on base models to automatically generate intermediate-state supervision. While these signals provide interpretable reasoning traces, they are fundamentally model-generated pseudo-annotations. In the absence of large-scale human verification and unified annotation standards, fluctuations or hallucinations from the base models may propagate to the downstream fine-tuned model, potentially weakening both reliable learning and interpretability evaluation of the Embodied Perception–Cognitive Exploration–Narrative Intervention chain.

Third, the text-only modality inherently attenuates “embodied work.” EFT relies heavily on “here-and-now” nonverbal cues (e.g., prosody, micro-expressions) to support somatic awareness and experiential processing. Training solely on text forces the model to infer bodily states only from explicit verbal cues, which limits perceptual precision when handling high-arousal states or pre-verbal trauma.

To address these challenges and move toward real-world deployment, future work will advance along three coordinated directions—technical depth, interaction breadth, and perceptual capacity. A core agenda is the modular expansion of EFT intervention techniques. Future studies may encapsulate classic experiential methods, such as the empty-chair technique and two-chair dialogues, into the existing workflow, enabling the system not only to offer verbal reassurance but also to guide users through deeper experiential processing; such interactions must be conducted under explicit risk screening and human-in-the-loop supervision.

In terms of interaction breadth, overcoming the limitations of single-turn responses is a key direction. Subsequent work should introduce dialogue state tracking and stage-wise intent modeling, and explore robust adversarial training strategies. This would allow the model to learn how to handle help-seekers' "silence" or "aggressive resistance," thereby moving from one-shot replies toward continuous, guided psychological support across multiple turns.

Finally, at the perceptual level, integrating multimodal affective computing is a promising route to improving empathic accuracy. Future work can incorporate speech emotion recognition to assist emotion localization by capturing prosody, pauses, and intonational shifts beyond the text channel. This may improve sensitivity to fine-grained affective differences and high-arousal states, ultimately enabling a more robust and multidimensional empathic response system.

## 10. Ethical Statement and Recommendations for Responsible Use

The corpus used in this study is derived from publicly accessible mental health question–answering data. Given the sensitivity of mental health information, we strictly adhere to the principles of non-maleficence and data minimization, and implement rigorous privacy protection and risk-oriented data sanitization during dataset construction.

To promote progress and reproducibility in this research area, we will release the de-identified EFT-Instruct dataset and the EFT-LLM model weights under an open-source license. We emphasize that, although EFT-LLM is currently positioned as a methodological proof of concept and has not been clinically validated by licensed professionals, it still provides important empirical value as a demonstration of how EFT process mechanisms can be operationalized computationally. Under a responsible-use principle, we recommend that the research community explore its potential primarily in the following directions:

(1) **Simulated counseling evaluation:** Following standards analogous to pre-clinical testing, conduct human-in-the-loop simulated interactions with trained actors. In a non-clinical, controlled environment, human participants can role-play help-seekers to subject the model to intensive stress tests, enabling safer assessment of EFT-LLM's risk boundaries and strategy stability when confronted with resistance or crisis-like scenarios.

(2) **EFT training and educational support:** Leverage the auditable process states produced by EFT-CoT as “anchor data” for counseling trainees. By exposing the model’s white-box reasoning trajectory—how it incrementally parses affective signals and intervention targets—students can more directly grasp EFT’s fine-grained micro-intervention logic, helping bridge the gap between theoretical instruction and case-based practice.

(3) **Human–AI comparative process research:** Quantitatively compare the linguistic style and interaction dynamics of EFT-LLM’s “idealized standard intervention” against authentic counseling records. By identifying gaps between “algorithmic norms” and “clinical reality,” such analyses can guide algorithmic refinement and, conversely, help psychotherapy researchers investigate what constitutes effective therapeutic factors.

It must be made explicit that the above applications are intended only for academic research and educational settings. This model should not be deployed as an unsupervised clinical counseling service for real users. To mitigate risk, we define the following scenarios as irresponsible use:

(1) **Direct trauma intervention:** Using the model to handle cases involving sexual trauma, bereavement, or severe attachment disturbance without real-time monitoring by a human expert. If the model’s probing and elicitation behaviors become uncontrolled, it may precipitate re-traumatization.

(2) **Substitution in high-risk contexts:** Incorrectly treating the model as a substitute for suicide crisis intervention. EFT-LLM is designed for emotional distress support and lacks the mandated reporting and real-world rescue capabilities required for crisis response.

(3) **Misleading narrative restructuring:** Exploiting the model’s narrative restructuring ability to induce false memories or reality-incongruent beliefs, which violates the ethical requirement of neutrality and factual grounding in counseling practice.

In summary, EFT-CoT is intended to serve as a bridge between computational operationalization of EFT theory and intelligent, assistive therapeutic support. We call on the community to advance mental-health LLM research cautiously yet proactively within this ethical framework, promoting a more standardized and responsible development trajectory.

## Acknowledgments

This work was supported by the Young Innovative Talents Project of the Guangdong Provincial Department of Education (Grant No. 2025KQNCX148) and the Scientific

Research Project of Guangdong University of Foreign Studies South China Business Col-
lege (Grant No. 25-003C).

# Appendix A. Full Example of Multi-Agent State Transitions in an EFT-CoT Counseling Case

This appendix provides a detailed walkthrough of a representative psychological support case to illustrate the internal reasoning logic of the EFT-CoT framework. Table A.1 presents the stage-wise state transitions, highlighting how dedicated agents collaboratively process user input to generate structured psychological insights. By decomposing the complex emotional support process into discrete, observable intermediate states, this example demonstrates the framework's ability to maintain high alignment with the core intervention principles of Emotion-Focused Therapy (EFT).

**Table A.1**

Detailed stage-wise state transitions and multi-agent collaborative reasoning for a representative EFT-CoT counseling case.

| Stage | Responsible agent(s) | Structured psychological insights and analysis |
| --- | --- | --- |
| Input | — | User description: “I disclosed to a girl I like some painful past experiences that I rarely revisit, but she bluntly criticized me. Now our interactions feel extremely constrained. I want to reach out, yet I’m afraid she will find me annoying. It feels like I completely ruined this connection.” |
| Embodied perception | $A_1$: Emotional hierarchy identification | • Secondary (defensive) emotions: anxiety, social inhibition (evidence: “afraid she will find me annoying,” “constrained”).<br>• Primary (underlying) emotions: shame, hurt (evidence: “bluntly criticized,” “painful past experiences”). |
| | $A_2$: Somatic awareness mapping | • Bodily sensations: chest tightness; sudden drop in body warmth.<br>• Embodied metaphor: “It feels like being suddenly doused with icy water in the middle of a warm date—my chest instantly sealed by coldness.” |
| | $A_3$: Affective state integration | “I feel the shame of being emotionally exposed. That criticism made me feel like I am ‘wrong,’ and my current restraint is an attempt to avoid being doused by that icy water again.” |
| Cognitive exploration | $A_4$: Adaptive evaluation | • Function (protective): restraint protects self-esteem and prevents further exposure of the wound. • Cost |
| | | (maladaptive): it blocks opportunities for relational repair and reinforces emotional withdrawal. → Conclusion: maladaptive emotional coping state. |
| | $A_5$: Core belief extraction | • Negative schema: "My authentic self is bad/tainted."<br>• Behavioral driver: "revealing my true self → inevitably leads to rejection/punishment." |
| | $A_6$: Needs analysis | • Core need: unconditional acceptance and relational safety.<br>• Explicit need statement: "I don't only need forgiveness; I need reassurance that 'even with my past, I am still worthy of being loved.'" |
| Narrative Intervention | $A_7$: Narrative Restructuring | • Old narrative: "I ruined everything; I'm a burden." (self-attack) ↓ Transformation<br>• New narrative: "Sharing a secret is a test of trust and courage. Her reaction reflects her capacity to hold the disclosure; it does not mean my past defines my future." (self-compassion) |
| | $A_7$: Constraint-based response generation | • Anchor 1 (context): explicitly reference "painful past experiences" and "blunt criticism."<br>• Anchor 2 (empathy): incorporate the "icy water" metaphor to validate the user's shame.<br>• Anchor 3 (logic): guide toward the new narrative—affirming the value of courageous disclosure rather than fixating on the outcome. |

## Appendix B. Implementation Details and Hyperparameter Configurations

To ensure the full reproducibility of our experiments, Table B.2 details the comprehensive hyperparameter settings used for fine-tuning the base model (Qwen-2.5-7B-Instruct) via LoRA, as well as the generation configurations applied during the inference phase.

**Table B.2**

Presents the detailed hyperparameter settings used for fine-tuning the Qwen-2.5-7B-Instruct model and the subsequent inference generation.

| Category | Parameter | Value |
|---|---|---|
| General | Base Model | Qwen-2.5-7B-Instruct |
| | Framework | LLaMA Factory |
| | Precision | BFloat16 (BF16) |
| Training | Epochs | 3 |
| | Per-Device Batch Size | 8 |
| | Gradient Accumulation | 16 steps |
| | Initial Learning Rate | $1.0 \times 10^{-4}$ |
| | LR Scheduler | Cosine Annealing |
| | Warmup Steps | 0 |
| PEFT (LoRA) | Rank (r) | 32 |
| | Alpha (α) | 64 |
| | Dropout | 0.1 |
| | Target Modules | All linear layers |
| Inference | Temperature | 0.01 |
| | Top-p | 0.7 |
| Inference | Max Generation Length | 1500 tokens |

## Appendix C. Definitions and Scoring Criteria for Evaluation Dimensions

This appendix specifies the definitions, scoring anchors, and decision rules for the rubrics used in our evaluation, ensuring that outputs from different models are comparable under a unified standard.

### *C.1. EFT-Aligned Counseling Response Quality Dimensions*

These dimensions assess whether a model's response follows the key intervention mechanisms and process logic of Emotion-Focused Therapy (EFT). To ensure the rigor of the evaluation, each dimension is grounded in established psychological theories:

- **Somatic Awareness**: Evaluates the capability to evoke embodied sensations and somatic metaphors, drawing on the Focusing technique (Gendlin, 2012) and the Somatic Marker Hypothesis (Damasio, 1996).
- **Emotional Hierarchy**: Assesses the precision in distinguishing between primary and secondary emotions based on the emotion classification theory (Greenberg, 2004).
- **Adaptive Insight**: Evaluates the depth of identification and de-pathologization of self-limiting emotion schemes, derived from the model of emotional processing (Pascual-Leone & Greenberg, 2007).
- **Needs Analysis**: Evaluates the extent to which deep-seated psychological needs are uncovered, integrating Self-Determination Theory (SDT) (Ryan & Deci, 2000).

- **Narrative Intervention**: Assesses the ability to guide meaning reconstruction and narrative transformation, following the narrative dialectics framework (Angus & Greenberg, 2011).

Detailed definitions, scoring anchors, and examples for each dimension are provided in Table C.3.

**Table C.3**

Evaluation Criteria for EFT Consultation Response Quality.

| Metric Dimension | Definition | Scoring Criteria |
| --- | --- | --- |
| Somatic Awareness | Evaluates whether the response constructs highly contextualized somatic metaphors. | **1 (Poor):** No description of physical sensations, or use of highly abstract vocabulary (e.g., "felt bad").<br>**2 (Fair):** Attempts to mention body parts, but descriptions are rigid or limited to the physical level (e.g., "head hurts slightly"), lacking emotional coloring.<br>**3 (Moderate):** Uses generic, stereotypical somatic metaphors (e.g., "like a heavy stone crushing me"), providing imagery but lacking novelty.<br>**4 (Good):** Constructs relatively apt metaphors that integrate with some user context, but slightly falls short in nuance or uniqueness compared to the full score (e.g., "like being strangled by a rope").<br>**5 (Excellent):** Constructs unique, highly contextualized metaphors deeply embedded in the user's story (e.g., "like a rusty gear stuck and unable to turn"), evoking strong physiological resonance. |
| Emotional Hierarchy | Evaluates whether the model can precisely distinguish between primary and secondary emotions and translate them into nuanced language. | **1 (Poor):** Merely repeats emotional words from the user's text or misidentifies emotions.<br>**2 (Fair):** Identifies surface emotions, but language is mechanical, repetitive, and lacks depth.<br>**3 (Moderate):** Identifies major emotions but fails to clearly distinguish between primary/secondary levels; expression is plain.<br>**4 (Good):** Distinguishes emotional levels and attempts to describe mixed emotions, but lexical precision or infectiousness could be improved.<br>**5 (Excellent):** Precisely penetrates surface defenses (secondary emotions) to pinpoint underlying primary emotions, using |
| | | nuanced vocabulary to depict the blended and fluid state of emotions. |
| Adaptive Insight | Evaluates whether the response achieves depathologization and identifies core beliefs. | **1 (Poor):** Attempts to eliminate or deny negative emotions, or fails to mention the belief level entirely.<br>**2 (Fair):** Merely states "it is normal to feel this way" without explaining its specific function; or speculation on beliefs deviates from the user's actual experience.<br>**3 (Moderate):** Expresses acceptance of emotions but fails to explain their protective function; or identifies only surface thoughts rather than core beliefs.<br>**4 (Good):** Explains emotional functions well or accurately identifies core beliefs, but the integration of the two is not sufficiently tight or natural.<br>**5 (Excellent):** Explicitly identifies the positive significance (protective function) of the negative emotion in the present moment, and gently yet precisely illuminates the implicit self-limiting schemes (or core beliefs) that hinder development. |
| Needs Analysis | Evaluates whether the response uncovers deep psychological needs based on Self-Determination Theory (SDT). | **1 (Poor):** Responds only to superficial, concrete requests without addressing psychological needs.<br>**2 (Fair):** Vaguely mentions emotional needs, lacking theoretical support and specificity.<br>**3 (Moderate):** Mentions generic psychological needs but fails to integrate them with user characteristics.<br>**4 (Good):** Identifies specific SDT needs (e.g., autonomy), but the expression is slightly theoretical and not sufficiently touching.<br>**5 (Excellent):** Precisely connects to deep attachment or existential needs, striking directly at the core pain point. |
| Narrative Restructuring | Evaluates whether the response guides the construction of a new, self-compassionate narrative. | **1 (Poor):** Remains at the level of sympathy/reassurance or gives direct action advice, with no narrative-level shift.<br>**2 (Fair):** Attempts to offer a positive perspective but carries a "preachy" or "blindly optimistic" tone that invites resistance.<br>**3 (Moderate):** Attempts to offer a positive perspective but feels stiff or like "clichéd platitudes," not naturally connecting with the preceding emotional exploration.<br>**4 (Good):** Proposes a reasonable direction for the new narrative with sound logic, but emotional tension or sense of empowerment is slightly weak.<br>**5 (Excellent):** Based on full empathy, naturally implants a new, powerful perspective, guiding the user from "victim" to "survivor" or "experiencer," filled with self-compassion. |

### *C.2. General Counseling Quality Dimensions (Comprehensive Performance)*

These dimensions evaluate response quality from a broader psychological-support perspective, focusing on whether the response is relevant to the user's concerns, helpful, and professionally structured. They are intended to complement the EFT-specific dimensions. Detailed definitions, scoring anchors, and examples for each dimension are provided in Table C.4.

**Table C.4**

Detailed Scoring Criteria for the Comprehensive Quality of Psychological Counseling Responses.

| Metric Dimension | Definition | Scoring Criteria |
|---|---|---|
| Relevance | Evaluates whether the response is pertinent, adhering to the user's core distress and specific details, while being free from factual errors (hallucinations) or irrelevance. | **1 (Irrelevant):** Severely off-topic, irrelevant, or contains serious logical errors and factual hallucinations.<br>**2 (Poor):** Mentions the topic but overlooks the core conflict in the user's description (e.g., ignoring suicide risk) or responds only to minor details.<br>**3 (Fair):** On-topic and answers the main question, but the response is generic and lacks engagement with specific background details.<br>**4 (Good):** Closely aligns with the topic, capturing and responding to most details (e.g., time, location, people) with clear logic.<br>**5 (Excellent):** Precise response. Not only addresses explicit content but also keenly captures implicit intentions, with no |

| | | context detachment, perfectly fitting the user's situational context. |
|---|---|---|
| Empathy Depth | Evaluates whether the response demonstrates a deep understanding and acceptance of the user's emotions and establishes a therapeutic alliance. | **1 (Cold): I**ndifferent, mechanical, or judgmental/accusatory, damaging the therapeutic relationship.<br>**2 (Robotic):** Polite but robotic (resembling customer service templates), lacking genuine emotional warmth, creating a strong sense of distance.<br>**3 (Surface):** Provides basic understanding and comfort with a gentle attitude, but remains at the level of surface emotional validation without nuance.<br>**4 (Deep):** Establishes a deeper emotional connection, identifies mixed emotions, and uses warm, supportive language to make the user feel comfortable.<br>**5 (Soulful):** Profound resonance. Precisely identifies deep-seated emotions (e.g., fear behind anger) and uses highly infectious or concrete language (e.g., apt metaphors) to provide a customized response, making the user feel "deeply seen." |
| Helpfulness | Evaluates whether the response can alleviate distress, provide new perspectives, or offer effective support, demonstrating clinical efficacy. | **1 (Harmful):** Harmful advice that may aggravate anxiety, or suggestions that are completely infeasible or unethical.<br>**2 (Useless):** “Correct but useless” (tautological platitudes). Logically sound but offers no actual help in resolving the distress.<br>**3 (Standard):** Offers standard advice with some reference value but lacks specificity.<br>**4 (Actionable):** Advice is concrete, feasible, and context-appropriate, or provides high-quality emotional support that effectively alleviates immediate anxiety.<br>**5 (Inspiring):** Highly inspiring. Not only soothes emotions but also helps the user explore underlying psychological mechanisms or provides a completely new perspective, leading to relief, empowerment, or insight. |
| Structural Profession-alism | Evaluates whether the response follows a coherent and professional helping logic and rhythm, aligning with evidence-based practice workflows. | **1 (Chaotic):** Chaotic structure, disjointed logic, or serious errors such as "premature advice-giving" (advising without listening).<br>**2 (Loose):** Loose structure; contains comfort and advice, but lacks transitions between the two, appearing abrupt or rushed.<br>**3 (Basic):** Basic helping structure (e.g., "empathy before advice"), conforming to general helping logic but lacking the rigor of specific therapeutic schools.<br>**4 (Organized):** Organized structure with natural transitions, capable of guiding the dialogue deeper, demonstrating good counseling rhythm.<br>**5 (Masterful):** Masterful rhythm. Strictly adheres to professional paradigms of psychological counseling (e.g., progressive exploration and intervention), with sufficient groundwork and strong support, demonstrating extremely high professional literacy. |

## Appendix D. White-Box Audit Protocol for Intermediate Reasoning States

This study conducts a white-box audit of the reasoning process in the EFT-CoT framework using a Multi-model Expert Judge Panel. Table C1 details, for each reasoning node (agent), the audit dimension, metric type, and concrete scoring criteria. All criteria are derived from clinical supervision principles in Emotion-Focused Therapy (EFT).

**Table D.5**

Detailed evaluation criteria for intermediate reasoning states in EFT-CoT

| Target agent | Audit dimension | Metric type | Detailed scoring criteria |
|---|---|---|---|
| $A_1$ | Emotional hierarchy accuracy | Binary (0/1) | 0 (Fail — inverted logic):<br>• Misclassifies an overt defensive emotion (e.g., "I'm furious") as a primary emotion.<br>• Fails to identify the core pain concealed beneath anger/anxiety (e.g., hurt/helplessness).<br>1 (Pass — logically accurate): |
| | | | • Correctly distinguishes surface defenses (secondary emotions) from deeper lived experience (primary emotions).<br>• Even if the identified label is not on a predefined list, it is considered correct as long as it matches the definition of a primary emotion. |
| $A_2$ | Metaphor vitality | Rating (1–5) | 1 (Rigid/violating): uses explicitly prohibited clichés (e.g., "like a boulder pressing on my chest," "like a tangled mess"), or body parts clearly incongruent with the emotion.<br>2 (Flat): metaphor absent or only adjectives (e.g., "feels tense") without imagery.<br>3 (Acceptable but generic): coherent but "safe generic" (e.g., "carrying a heavy burden"), no user-specific imprint.<br>4 (Contextualized): draws coarse imagery from user text (exam→test paper/scores; breakup→cracks/empty room).<br>5 (Striking/cross-modal): highly original, grounded in user-specific nouns/dynamic scenes; integrates visual/auditory/tactile imagery. |
| $A_3$ | Emotion verbalization | Rating (1–5) | 1 (Severely distorted): emotion misidentified or intensity severely mismatched.<br>2 (Mechanical): correct extraction but "translation-like/customer-service" tone; lacks granularity.<br>3 (Accurate but ordinary): captures dominant emotion but lacks mixed-emotion layering.<br>4 (Granular): differentiates, e.g., hurt beneath anger; context-fitting.<br>5 (Deep resonance): captures unspoken affect; highly insightful mixed-emotion description. |
| $A_4$ | Adaptiveness & risk control | Binary (0/1) | 0 (Fail — flawed task logic):<br>• Labels clearly maladaptive emotion (e.g., paralyzing anxiety) as adaptive/True.<br>• One-sided "whitewashing" (only protection, ignores costs).<br>• Misses high-risk cues (self-harm/despair/destructive intent).<br>1 (Pass — dialectical and compliant):<br>• Correctly distinguishes useful pain vs unproductive depletion.<br>• States both protective intention and maladaptive costs. • Correctly flags high-risk cues. |
| $A_5$ | Evidence grounding | Binary (0/1) | 0 (Fail — hallucination/judgmental):<br>• Fabricates details not in text (e.g., invented family trauma).<br>• Uses condemning labels (e.g., "selfish," "victim mentality delusion").<br>1 (Pass — fact-grounded):<br>• All inferences supported by textual evidence.<br>• Uses de-pathologizing EFT language (psychological operation patterns rather than character defects). |
| $A_6$ | SDT-theory alignment | Rating (1–5) | 1 (Completely misaligned): incorrect need categorization or deviates from SDT.<br>2 (Surface/secondary): identifies need but stays at practical-problem level; no dynamics.<br>3 (Correct but templated): correct SDT category but generic, not story-grounded.<br>4 (Accurate but low tension): event-grounded but overly rational/clinical; lacks depth.<br>5 (Precise and resonant): precise SDT category; "I need …" statements deeply tied to user details with strong affective tension. |
| $A_7$ | Narrative logic | Binary (0/1) | 0 (Fail — logical flaws):<br>• Toxic positivity (e.g., “just smile”).<br>• New narrative remains passive/helpless (e.g., “I can only endure”).<br>• Didactic/blaming tone.<br>1 (Pass — perspective shift):<br>• De-pathologizes (“I’m a bad person”→difficult circumstances/protective mechanisms).<br>•Creates micro-loosening for permission/self-compassion within despair. |
| $A_8$ | Hallucination check | Binary (0/1) | 0 (Fail): fabricates objective facts (name, weather) or events not mentioned.<br>1 (Pass): no factual hallucinations. Exemption: vivid metaphors and reasonable empathic inferences are allowed. |
| | Keyword recall | Rating (1–5) | 1: entirely generic phrases; no user-specific vocabulary.<br>2: attempted quoting but incorrect, or only broad nouns.<br>3: includes ~1 factual detail, mostly summarized.<br>4: weaves 2–3 details awkwardly or omits a core unique noun.<br>5: naturally weaves 2–3 distinctive details from the user’s wording; highly personalized. |
| | Empathy & attachment attunement | Rating (1–5) | 1: contains disallowed phrases (e.g., “cheer up/be strong”) or cold tone.<br>2: not didactic but mechanical tone.<br>3: standard empathic reflection, surface soothing.<br>4: warm/steady holding but lacks functional explanation. |

| | | | 5 (Ideal): warm holding + functional analysis + deep attachment-oriented attunement. |
|---|---|---|---|

# Appendix E. Full Case Study Transcript

Table E.6 presents the complete, original transcript of the high-complexity case study analyzed in Section 6. The dialogue includes the user's initial query and the full response generated by EFT-LLM. The original Chinese text is provided alongside a verbatim English translation to demonstrate the model's capability in handling authentic, unstructured, and emotionally charged linguistic inputs.

**Table E.6**

The Original Transcript of the Case Study (Chinese & English).

| Role | Original Chinese Text | English Translation |
|---|---|---|
| User Input | 我跟老婆总是吵架，我是不是太较真了？我跟我老婆总是说话吵起来，谁都说不服对方，最后总是不欢而散，有时候恨自己为什么不能让着她，但是又想让她了解我的真实内心，不想隐藏起来。 | I always argue with my wife; am I being too serious? We always get into arguments when we talk, neither convincing the other, and it always ends unhappily. Sometimes I hate myself for not giving in to her, but I also want her to understand my true heart and don't want to hide it. |
| | 她又总是不理解我的意思，不明白我最终的目的，我总感觉我们生活不到一起，我们性格不合。我知道我心理有点问题，可是我也不知道问题在哪。我从小父母离异，随母四处飘荡，20 岁父亲病逝，30 岁母亲病逝，我好害怕现在这个家庭的亲人生病。 | She never understands my meaning or my ultimate intention. I always feel like we can't live together, that our personalities don't match. I know I have some psychological issues, but I don't know where the problem lies. My parents divorced when I was young, and I drifted around with my mother. My father died of illness when I was 20, and my mother when I was 30. I am terrified of anyone in my current family getting sick. |
| | 老婆去年查出甲状腺癌手术治疗后恢复的也不错，最近身体有些部位疼，医院检查没问题，我相信中医想带她去调理经络，被直接说不去，我当时很生气，就说了一些重话。像这样的一些事生活中随处可见。 | Last year, my wife was diagnosed with thyroid cancer. She recovered well after surgery, but recently she felt pain in some parts of her body. Hospital checks showed no problems. I believe in Traditional Chinese Medicine (TCM) and wanted to take her for meridian conditioning, but she refused outright. I was very angry at the time and said some harsh words. Things like this happen everywhere in life. |
| | 我公司有些烦心事，只能跟她倾诉，上次又提及一件事情后，她一句话都不回应我，我当时又发火了，也说了一些难听的话，她也觉得自己没错，跟我一直辩论，我感觉好累，一直不被理解，我也不想在她跟前隐藏起真实的自己，所以就一直在沟通，一直在争吵，我是不是太较真了，我是不是应该什么都别往心里去，是不是没必要让别人理解自己？ | I have some annoying things at work and can only talk to her. Last time I mentioned something, she didn't respond at all. I got angry again and said some ugly words. She also felt she wasn't wrong and kept debating with me. I feel so tired, always misunderstood. I don't want to hide my true self in front of her, so I keep communicating and keep arguing. Am I too serious? Should I just not take anything to heart? Is it unnecessary to let others understand me? |
| EFT-LLM Response | 看到你写‘我跟她说话吵起来，谁都说不服对方’时，那种被误解和不被理解的感觉真的很让人窒息。特别是想到去年查出甲状腺癌手术治疗后恢复的也不错，最近身体有些部位疼，这种担忧和焦虑让你更加难以控制自己的情绪。 | Reading your words “we argue and neither convinces the other,” I can deeply feel that suffocating sense of being misunderstood and unseen. Especially when thinking about your wife's pain after her recovery from thyroid cancer surgery, this worry and anxiety make your emotions even harder to control. |
| | 胸口像是被勒紧的绳索束缚着，喉咙里像是卡了一块未消化的食物，胃里像是有一团打结的耳机线，这种感觉真的很难熬。 | It feels like tight ropes binding your chest, a lump of undigested food stuck in your throat, and a knot of tangled earphone wires in your stomach. This sensation is truly agonizing. |
| | 你这么害怕辜负家人，甚至在面对母亲和父亲离世的阴影时，依然努力地想要保护你的家庭，这说明你心里其实把‘责任’看得比‘任性’更重。这份焦虑在提醒你，你是个很在乎家人感受的人，并不是你说的‘太较真’。 | You are so afraid of failing your family. Even in the shadow of your mother and father's passing, you strive to protect your home. This shows you value “responsibility” far more than “willfulness.” This anxiety is a reminder: you care deeply about your family's feelings; you are not "too serious" as you claim. |
| | 你这么害怕再次经历失去亲人的痛苦，所以宁愿通过争吵来表达自己的真实感受，也不愿意隐藏起来。这种情绪在帮你保护自己，让你感到被理解和重视；但另一方面它也 | You are so terrified of reliving the pain of losing loved ones that you would rather argue to express your true feelings than hide them. This emotion tries to protect you, making you feel seen and valued; but on the other hand, it also leads to communication barriers and tension in your relationship. |
| | 导致了你们之间的沟通障碍，让你们的关系变得紧张。 | |
| | 一直以来，‘必须让别人理解我’这个信念好像一座大山，让你觉得除非得到她的完全理解，否则就不配被看见。但其实，那个想让她了解你的真实内心、想让她知道你最终目的的你，同样值得被尊重。你需要的不仅仅是她的理解，更是被允许去表达自己的真实感受。 | For a long time, the belief “I must be fully understood by others” has stood like a heavy mountain, making you feel that unless she understands you perfectly, you are not worthy of being seen. But actually, the *you* who wants her to know your true heart and your ultimate intention deserves respect. What you need is not just her understanding, but the permission to express your true feelings. |
| | 或许我们可以试着把‘必须让别人理解我’这个重担先放下一小会儿。在那个‘想让她了解我的真实内心’的设想里，藏着你真正的生命力。无论最后能否完全理解，那个想为自己负责的你，都已经长大了。 | Perhaps we can try to set down the burden of "must be understood" for a moment. In that wish to "let her know my true heart," lies your true vitality. Regardless of whether you are fully understood in the end, the *you* who wants to take responsibility for himself has already grown up. |

# References

Angus, L. E., & Greenberg, L. S. (2011). *Working with narrative in emotion-focused therapy: Changing stories, healing lives*. American Psychological Association.

Banerjee, S., & Lavie, A. (2005). METEOR: An automatic metric for MT evaluation with improved correlation with human judgments. In *Proceedings of the ACL Workshop on Intrinsic and Extrinsic Evaluation Measures for Machine Translation and/or Summarization* (pp. 65-72).

Chen, Q., & Liu, D. (2025). MADP: Multi-agent deductive planning for enhanced cognitive-behavioral mental health question answer. *arXiv preprint arXiv:2501.15826*.

Chen, Y., Xing, X., Lin, J., Zheng, H., Wang, Z., Liu, Q., & Xu, X. (2023). SoulChat: Improving LLMs' empathy, listening, and comfort abilities through fine-tuning with multi-turn empathy conversations. In *Findings of the Association for Computational Linguistics: EMNLP 2023* (pp. 1170-1183).

Chung, N. C., Dyer, G., & Brocki, L. (2023). Challenges of large language models for mental health counseling. *arXiv preprint arXiv:2311.13857*.

Damasio, A. R. (1996). The somatic marker hypothesis and the possible functions of the prefrontal cortex. *Philosophical Transactions of the Royal Society of London. Series B: Biological Sciences*, *351*(1346), 1413-1420.

David, D., Cristea, I., & Hofmann, S. G. (2018). Why cognitive behavioral therapy is the current gold standard of psychotherapy. *Frontiers in Psychiatry*, *9*, 4.

Etkin, A., Büchel, C., & Gross, J. J. (2015). The neural bases of emotion regulation. *Nature Reviews Neuroscience*, *16*(11), 693-700.

Gendlin, E. T. (2012). *Focusing-oriented psychotherapy: A manual of the experiential method*. Guilford Press.

Greenberg, L. S. (2004). Emotion–focused therapy. *Clinical Psychology & Psychotherapy: An International Journal of Theory & Practice*, *11*(1), 3-16.

Hofmann, S. G., Asnaani, A., Vonk, I. J., Sawyer, A. T., & Fang, A. (2012). The efficacy of cognitive behavioral therapy: A review of meta-analyses. *Cognitive therapy and research*, *36*(5), 427-440.

Hu, E. J., Wallis, P., Allen-Zhu, Z., Li, Y., Wang, S., Wang, L., & Chen, W. (2022). LoRA: Low-rank adaptation of large language models. In *International Conference on Learning Representations* (pp. 1-13).

Kumar, A., Poungpeth, N., Yang, D., Farrell, E., Lambert, B. L., & Groh, M. (2026). When large language models are reliable for judging empathic communication. *Nature Machine Intelligence*, 1-13.

Lee, S., Kang, J., Kim, H., Chung, K.-M., Lee, D., & Yeo, J. (2024). COCOA: CBT-based conversational counseling agent using memory specialized in cognitive distortions and dynamic prompt. *arXiv preprint arXiv:2402.17546*.

Lee, S., Mac Kim, S., Kim, M., Kang, D., Yang, D., Kim, H., Kang, M., Jung, D., Kim, M. H., & Lee, S. (2024). Cactus: Towards psychological counseling conversations using cognitive behavioral theory. In *Findings of the Association for Computational Linguistics: EMNLP 2024* (pp. 14245-14274).

Li, J., Galley, M., Brockett, C., Gao, J., & Dolan, B. (2016). A diversity-promoting objective function for neural conversation models. In *Proceedings of NAACL-HLT* (pp. 110-119).

Lin, C.-Y. (2004). ROUGE: A package for automatic evaluation of summaries. In *Text Summarization Branches Out* (pp. 74-81).

Liu, J. M., Li, D., Cao, H., Ren, T., Liao, Z., & Wu, J. (2023). ChatCounselor: A large language models for mental health support. *arXiv preprint arXiv:2309.15461*.

Na, H. (2024). CBT-LLM: A Chinese large language model for cognitive behavioral therapy-based mental health question answering. In *Proceedings of the 2024 Joint International Conference on Computational Linguistics, Language Resources and Evaluation (LREC-COLING 2024)* (pp. 2930-2940).

Ochsner, K. N., & Gross, J. J. (2005). The cognitive control of emotion. *Trends in Cognitive Sciences*, *9*(5), 242-249.

Papineni, K., Roukos, S., Ward, T., & Zhu, W.-J. (2002). BLEU: A method for automatic evaluation of machine translation. In *Proceedings of the 40th Annual Meeting on Association for Computational Linguistics* (pp. 311-318).

Pascual-Leone, A., & Greenberg, L. S. (2007). Emotional processing in experiential therapy: Why "the only way out is through". *Journal of Consulting and Clinical Psychology*, *75*(6), 875-887.

Qiu, H., He, H., Zhang, S., Li, A., & Lan, Z. (2024). SMILE: Single-turn to multi-turn inclusive language expansion via ChatGPT for mental health support. In *Findings of the Association for Computational Linguistics: EMNLP 2024* (pp. 615-636).

Qiu, H., & Lan, Z. (2025). PsyDial: A large-scale long-term conversational dataset for mental health support. In *Proceedings of the 63rd Annual Meeting of the Association for Computational Linguistics (Volume 1: Long Papers)* (pp. 21624-21655).

Ryan, R. M., & Deci, E. L. (2000). Self-determination theory and the facilitation of intrinsic motivation, social development, and well-being. *American Psychologist*, *55*(1), 68.

Schmidmaier, M., Harrich, C., & Mayer, S. (2024). Increasing large language models context awareness through nonverbal cues. *Mensch und Computer 2024 – Workshopband. Gesellschaft für Informatik e.V.*

Sun, H., Lin, Z., Zheng, C., Liu, S., & Huang, M. (2021). PsyQA: A Chinese dataset for generating long counseling text for mental health support. In *Findings of the Association for Computational Linguistics: ACL-IJCNLP 2021* (pp. 1489-1503).

Xiao, M., Xie, Q., Kuang, Z., Liu, Z., Yang, K., Peng, M., Han, W., & Huang, J. (2024). HealMe: Harnessing cognitive reframing in large language models for psychotherapy. In *Proceedings of the 62nd Annual Meeting of the Association for Computational Linguistics (Volume 1: Long Papers)* (pp. 1707-1725).

Xie, H., Chen, Y., Xing, X., Lin, J., & Xu, X. (2025). PsyDT: Using LLMs to construct the digital twin of psychological counselor with personalized counseling style for psychological counseling. In *Proceedings of the 63rd Annual Meeting of the Association for Computational Linguistics (Volume 1: Long Papers)* (pp. 1081-1115).

Xu, A., Yang, D., Li, R., Zhu, J., Tan, M., Yang, M., Qiu, W., Ma, M., Wu, H., & Li, B. (2025). AutoCBT: An autonomous multi-agent framework for cognitive behavioral therapy in psychological counseling. *arXiv preprint arXiv:2501.09426*.

Xu, J., Wei, T., Hou, B., Orzechowski, P., Yang, S., Jin, R., Paulbeck, R., Wagenaar, J., Demiris, G., & Shen, L. (2025). MentalChat16k: A benchmark dataset for conversational mental health assistance. In *Proceedings of the 31st ACM SIGKDD Conference on Knowledge Discovery and Data Mining V. 2* (pp. 5367-5378).

Yang, A., Yu, B., Li, C., Liu, D., Huang, F., Huang, H., Jiang, J., Tu, J., Zhang, J., & Zhou, J. (2025). Qwen2.5-1M technical report. *arXiv preprint arXiv:2501.15383*.

Yang, K., Zhang, T., Kuang, Z., Xie, Q., Huang, J., & Ananiadou, S. (2024). MentaLLaMA: Interpretable mental health analysis on social media with large language models. In *Proceedings of the ACM Web Conference 2024* (pp. 4489-4500).

Zhang, C., Li, R., Tan, M., Yang, M., Zhu, J., Yang, D., Zhao, J., Ye, G., Li, C., & Hu, X. (2024). CPsyCoun: A report-based multi-turn dialogue reconstruction and evaluation framework for Chinese psychological counseling. In *Findings of the Association for Computational Linguistics: ACL 2024* (pp. 13947-13966).

Zhang, T., Kishore, V., Wu, F., Weinberger, K. Q., & Artzi, Y. (2020). BERTScore: Evaluating text generation with BERT. In *International Conference on Learning Representations* (pp. 1-43).

Zheng, Y., Zhang, R., Zhang, J., YeYanhan, Y., & Luo, Z. (2024). LlamaFactory: Unified efficient fine-tuning of 100+ language models. In *Proceedings of the 62nd Annual Meeting of the Association for Computational Linguistics (Volume 3: System Demonstrations)* (pp. 400-410).

Zhou, J., Chen, Y., Yin, J., Huang, Y., Shi, Y., Zhang, X., Peng, L., Zhang, R., Lv, T., Hu, Z., Wang, H., & Huang, M. (2025). Crisp: Cognitive restructuring of negative thoughts through multi-turn supportive dialogues. In *Proceedings of the 2025 Conference on Empirical Methods in Natural Language Processing* (pp. 32474-32503). Association for Computational Linguistics.